\def\year{2021}\relax
\documentclass[letterpaper]{article} 
\usepackage{aaai21}  
\usepackage{times}  
\usepackage{helvet} 
\usepackage{courier}  
\usepackage[hyphens]{url}  
\usepackage{graphicx} 
\usepackage{natbib}  
\usepackage{caption} 
\usepackage{hyperref}       
\usepackage{booktabs}       
\usepackage{amsfonts}       
\usepackage{nicefrac}       
\usepackage{microtype}      
\usepackage{natbib}
\usepackage{amsmath}
\usepackage{enumitem}
\usepackage{amssymb}
\DeclareMathOperator*{\argmax}{arg\,max}
\usepackage{bbm}
\usepackage{adjustbox}
\usepackage{caption}
\usepackage{subcaption}
\usepackage{verbatimbox}
\usepackage{multirow}
\usepackage{color, colortbl}
\usepackage{stfloats}
\title{Neural Analogical Matching}
\author{
 Maxwell Crouse\textsuperscript{\rm 1}\thanks{Correspondence to \texttt{mvcrouse@u.northwestern.edu}, code available at \url{https://github.com/mvcrouse/NeuralAnalogy}.
 },
 Constantine Nakos\textsuperscript{\rm 1},
 Ibrahim Abdelaziz\textsuperscript{\rm 2},
 Kenneth Forbus\textsuperscript{\rm 1} \\
}
\affiliations{
 \textsuperscript{\rm 1}Qualitative Reasoning Group, Northwestern University \\
 \textsuperscript{\rm 2}IBM Research, IBM T.J. Watson Research Center \\
 \{mvcrouse, cnakos\}@u.northwestern.edu, ibrahim.abdelaziz1@ibm.com, forbus@northwestern.edu
}

\begin{document}

\maketitle

\begin{abstract}
Analogy is core to human cognition. It allows us to solve problems based on prior experience, it governs the way we conceptualize new information, and it even influences our visual perception. The importance of analogy to humans has made it an active area of research in the broader field of artificial intelligence, resulting in data-efficient models that learn and reason in human-like ways. 
While cognitive perspectives of analogy and deep learning have generally been studied independently of one another, the integration of the two lines of research is a promising step towards more robust and efficient learning techniques.
As part of a growing body of research on such an integration, we introduce the Analogical Matching Network: a neural architecture that learns to produce analogies between structured, symbolic representations that are largely consistent with the principles of Structure-Mapping Theory.
\end{abstract}

\section{Introduction}
\label{sec:introduction}

Analogical reasoning is a form of inductive reasoning that cognitive scientists consider to be one of the cornerstones of human intelligence \cite{gentner2003we,hofstadter2001analogy,hofstadter1995fluid}.  Analogy shows up at nearly every level of human cognition, from low-level visual processing \cite{sagi2012difference} to abstract conceptual change \cite{gentner1997analogical}.  Problem solving using analogy is common, with past solutions forming the basis for dealing with new problems \cite{holyoak1984development,novick1988analogical}. Analogy also facilitates learning and understanding by allowing people to generalize specific situations into increasingly abstract schemas \cite{gick1983schema}.

Many different theories have been proposed for how humans perform analogy \cite{mitchell1993analogy,chalmers1992high,gentner1983structure,holyoak1996mental}. One of the most influential theories is Structure-Mapping Theory (SMT) \cite{gentner1983structure}, which posits that analogy involves the alignment of structured representations of objects or situations subject to certain constraints. Key characteristics of SMT are its use of symbolic representations and its emphasis on relational structure, which allow the same principles to apply to a wide variety of domains.

Until now, the symbolic, structured nature of SMT has made it a poor fit for deep learning. The representations produced by deep learning techniques are incompatible with off-the-shelf SMT implementations like the Structure-Mapping Engine (SME) \cite{falkenhainer1989structure,forbus2017extending}, while the symbolic graphs that SMT assumes as input are challenging to encode with traditional neural methods. In this work, we describe how recent advances in graph representation learning can be leveraged to create deep learning systems that can learn to produce structural analogies consistent with SMT.

\subsubsection{Contributions:} We introduce the Analogical Matching Network (AMN), a neural architecture that learns to produce analogies between symbolic representations. 
AMN is trained on purely synthetic data and is demonstrated over a diverse set of analogy problems drawn from structure-mapping literature to produce outputs that are largely consistent with SMT. With AMN, we aim to push the boundaries of deep learning and extend them to an important area of human cognition; in particular, by showing how to design a deep learning system that conforms to a cognitive theory of analogical reasoning. It is our hope that future generations of neural architectures can reap the same benefits from analogy that symbolic reasoning systems and humans currently do.







\section{Related Work}

Many different computational models of analogy have been proposed \cite{mitchell1993analogy,holyoak1989analogical,o1999computability,forbus2017extending}, each instantiating a different cognitive theory of analogy. The differences between them are compounded by the computational costs of analogical reasoning, a provably NP-Hard problem \cite{veale1997competence}. While these computational models are often used to test cognitive theories of human behavior, they are also useful tools for applied tasks. For instance, the Structure-Mapping Engine (SME) has been used in question-answering \cite{ribeiro2013predicting}, computer vision \cite{chen2019human}, and machine reasoning \cite{klenk2005solving}.

Many of the early approaches to analogy were connectionist \cite{gentner1993analogy}. The STAR architecture of \cite{halford1994connectionist} used tensor product representations of structured data to perform simple analogies of the form $R(x, y) \Rightarrow S(f(x), f(y))$.
Drama \cite{eliasmith2001integrating} was an implementation of the multi-constraint theory of analogy \cite{holyoak1996mental} that used holographic representations similar to tensor products to embed structure. 
LISA \cite{hummel1997distributed,hummel2005relational} was a hybrid symbolic connectionist approach to analogy. It staged the mapping process temporally, generating mappings from elements that were activated at the same time.

Cognitive perspectives of analogy have gone relatively unexplored in deep learning research, with only a few recent works that address them \cite{hill2019learning,zhang2019raven,lu2019seeing}. Most prior deep learning works have considered analogies involving perceptual data \cite{mikolov2013linguistic,reed2015deep,bojanowski2017enriching,zhou2019learning,benaim2020structural}. Such problems differ from those seen in the structure-mapping literature in that they typically do not require explicit graph matching and they involve only one relation which is unobserved. 

Our approach is conceptually related to recent work on neural graph matching \cite{emami2018learning,georgiev2020neural,wang2019learning}. Such works generally focus on finding unconstrained maximum weight matchings and often interleave their networks with hard-coded algorithms (e.g., \cite{emami2018learning} applies the Hungarian algorithm to coerce its outputs into a permutation matrix). These considerations make them less applicable here, as 1) SMT is subject to unique constraints that make standard bipartite matching techniques insufficient and 2) we wish to explore the extent to which SMT is purely learnable.

\begin{figure*}[t]
\centering
    \footnotesize
    \begin{tabular}{l l | l l}
        \toprule
        \texttt{[1]} & \texttt{nucleus} & \texttt{[8]} & \texttt{sun}\\
         \texttt{[2]} & \texttt{electron} & \texttt{[9]} & \texttt{planet}\\
         \texttt{[3]} & \texttt{MASS([1])} & \texttt{[10]} & \texttt{MASS([8])}\\
        \texttt{[4]} & \texttt{MASS([2])} & \texttt{[11]} & \texttt{MASS([9])} \\
         \texttt{[5]} & \texttt{ATTRACTS([1]], [2])} & \texttt{[12]} & \texttt{TEMPERATURE([8]])}\\
         \texttt{[6]} & \texttt{REVOLVES-AROUND([2], [1])} & \texttt{[13]} & \texttt{TEMPERATURE([9]])} \\
        \texttt{[7]} & \texttt{GREATER([3], [4])} & \texttt{[14]} & \texttt{REVOLVES-AROUND([9], [8])} \\
         & & \texttt{[15]} & \texttt{GREATER([10], [11])}\\
         & & \texttt{[16]} & \texttt{GREATER([12], [13])} \\
         & & \texttt{[17]} & \texttt{ATTRACTS([9], [8])} \\
         & & \texttt{[18]} & \texttt{CAUSES(AND([15], [17]), [14])} \\
         & & \texttt{[19]} & \texttt{YELLOW([8]])} \\
        \bottomrule
    \end{tabular}
    \includegraphics[width=0.8\textwidth]{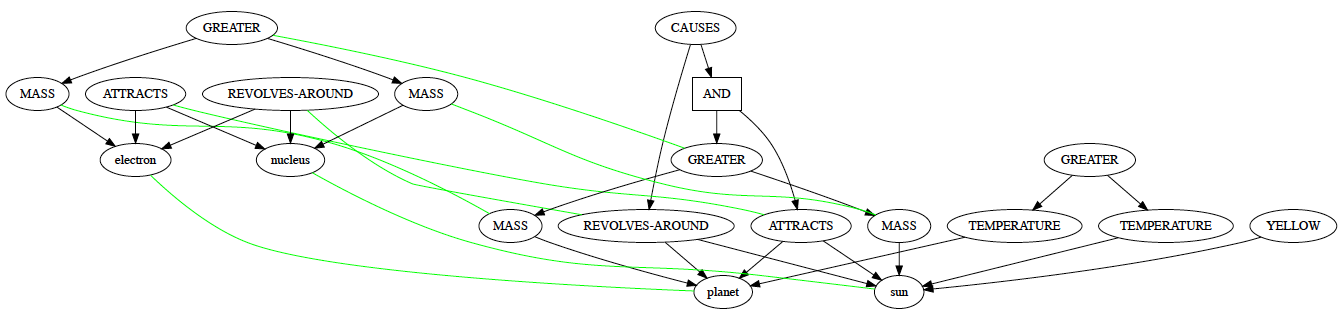}
    \caption{Relational and graph representations for models of the atom (left) and Solar System (right). Light green edges indicate the set of correspondences between the two graphs. }
    \label{fig:rel_repr}
\end{figure*}

\section{Structure-Mapping Theory}
\label{sec:smt}

In Structure-Mapping Theory (SMT) \cite{gentner1983structure}, analogy centers around the structural alignment of \textit{relational representations} (see Figure \ref{fig:rel_repr}). A relational representation is a set of logical expressions constructed from entities (e.g., \texttt{sun}), attributes (e.g., \texttt{YELLOW}), functions (e.g., \texttt{TEMPERATURE}), and relations (e.g., \texttt{GREATER}).
Structural alignment is the process of producing a \textit{mapping} between two relational representations (referred to as the \textit{base} and \textit{target}). A mapping is a triple $\big< M, C, S \big>$, where $M$ is a set of \textit{correspondences} between the base and target, $C$ is a set of \textit{candidate inferences} (i.e., inferences about the target that can be made from the structure of the base), and $S$ is a \textit{structural evaluation score} that measures the quality of $M$. Correspondences are pairs of elements between the base and target (i.e., expressions or entities) that are identified as matching with one another. While entities can be matched together irrespective of their labels, there are more rigorous criteria for matching expressions. SMT asserts that matches should satisfy the following:
\begin{enumerate}[leftmargin=*]
    \item \textit{One-to-One}: Each element of the base and target can be a part of \textit{at most} one correspondence.
    \item \textit{Parallel Connectivity}: Two expressions can be in a correspondence with each other only if their arguments are also in correspondences with each other.
    \item \textit{Tiered Identicality}: Relations of expressions in a correspondence must match identically, but functions need not if their correspondence supports parallel connectivity.
    \item \textit{Systematicity}: Preference should be given to mappings with more deeply nested expressions.
\end{enumerate}

To understand these properties, we use a classic analogy (see Figure \ref{fig:rel_repr}) from \cite{gentner1983structure,falkenhainer1989structure}, which draws an analogy between the Solar System and the Rutherford model of the atom. A set of correspondences $M$ between the base (Solar System) and target (Rutherford atom) is a set of pairs of elements from both sets, e.g., $\{\big< \texttt{[1]}, \texttt{[8]} \big>, \big< \texttt{[2]}, \texttt{[9]} \big> \}$. The one-to-one constraint restricts each element to be a member of at most one correspondence. Thus, if $\big<\texttt{[7]}, \texttt{[15]}\big>$ was a member of $M$, then $\big<\texttt{[7]}, \texttt{[16]}\big>$ could not be added to $M$. Parallel connectivity enforces correspondence between arguments if the parents are in correspondence. In this example, if $\big<\texttt{[7]}, \texttt{[15]}\big>$ was a member of $M$, then both $\big<\texttt{[3]}, \texttt{[10]}\big>$ and $\big<\texttt{[4]}, \texttt{[11]}\big>$ would need to be members of $M$. Parallel connectivity also respects argument order when dealing with ordered relations. Tiered identicality is not relevant in this example; however, if \texttt{[10]} used the label \texttt{WEIGHT} instead of \texttt{MASS}, tiered identicality could be used to match \texttt{[3]} and \texttt{[10]}, since such a correspondence would allow for a match between their parents. The last property, systematicity, results in larger correspondence sets being preferred over smaller ones. Note that the singleton set $\{ \big<\texttt{[1]}, \texttt{[8]}\big> \}$ satisfies SMT's constraints, but it is clearly not useful by itself. Systematicity captures the natural preference for larger, more interesting matches.

Candidate inferences are statements from the base that are projected into the target to fill in missing structure \cite{bowdle1997informativity,gentner1998analogy}. Given a set of correspondences $M$, candidate inferences are created from statements in the base that are supported by expressions in $M$ but are not part of $M$ themselves. In Figure \ref{fig:rel_repr}, one candidate inference would be \texttt{CAUSES(AND([7],[5]),[6])}, derived from \texttt{[18]} by substituting its arguments with the expressions they correspond to in the target.
In this work, we adopt SME's default criteria for computing candidate inferences. Valid candidate inferences are all statements that have \emph{some} dependency that is included in the correspondences or an ancestor that is a candidate inference (e.g., an expression whose parent has arguments in the correspondences).

\begin{figure*}[t]
    \centering
    \includegraphics[width=0.8\textwidth]{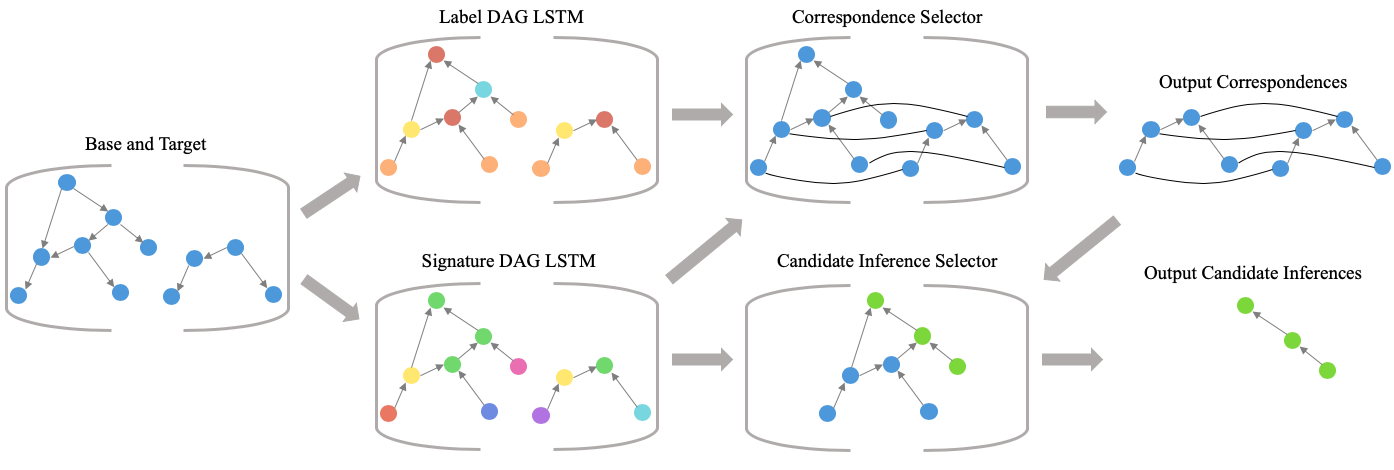}
    \caption{An overview of the model pipeline}
    \label{fig:model_overview}
\end{figure*}

The concepts above carry over naturally into graph-theoretic notions. The base and target are considered semi-ordered directed-acyclic graphs (DAGs) $G_B = \big<V_B, E_B\big>$ and $G_T = \big<V_T, E_T\big>$, where $V_B$ and $V_T$ are sets of nodes and $E_B$ and $E_T$ are sets of edges. Each node corresponds to some expression and has a label given by its relation, function, attribute, or entity name. 
Structural alignment is the process of finding a maximum weight bipartite matching $M \subseteq V_B \times V_T$, where 
$M$ satisfies the pairwise-disjunctive constraints imposed by parallel connectivity. Finding candidate inferences is then determining the subset of nodes from $V_B \setminus \{ b_i : \big<b_i, t_j\big> \in M\}$ with support in $M$.

\section{Model}

\subsection{Model Components}
\label{sec:model_comps}


Given a base $G_B = \big<V_B, E_B\big>$ and target $G_T = \big<V_T, E_T\big>$, AMN produces a set of correspondences $M \subseteq V_B \times V_T$ and a set of candidate inferences $I \in V_B \setminus \{b_i : \big<b_i, t_j\big> \in M\}$.
A key design choice of this work was to avoid using rules or architectures that force particular outputs whenever possible. AMN is \emph{not} forced to output correspondences that satisfy the constraints of SMT; instead, conformance with SMT is reinforced through performance on training data. Our architecture uses Transformers \cite{vaswani2017attention} and pointer networks \cite{vinyals2015pointer} and takes inspiration from the work of \cite{kool2018attention}. A high-level overview is given in Figure \ref{fig:model_overview}, which shows how each of the three main components (graph embedding, correspondence selection, and candidate inference selection) interact with one another.

\subsubsection{Representing Structure:} 
\begin{figure*}[t]
\renewcommand*{\arraystretch}{1.}
\centering
\includegraphics[width=\textwidth]{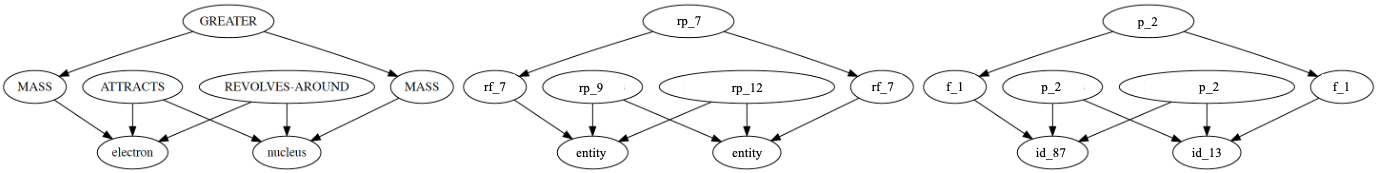}
\caption{Original graph (left), its label graph (middle), and its signature graph (right)}
\label{fig:three_reps}
\end{figure*}

When embedding the nodes of $G_B$ and $G_T$, there are representational concerns to keep in mind. First, as matching should be done on the basis of structure, the labels of entities should not be taken into account during the alignment process. Second, because SMT's constraints require AMN to be able to recognize when a node is part of multiple correspondences, AMN should maintain distinguishable representations for distinct nodes, even if those nodes have the same labels. Last, the architecture should not be vocabulary dependent, i.e., AMN should generalize to symbols it has never seen before. To achieve each of these, AMN first parses the original input into two separate graphs, a \textit{label graph} and a \textit{signature graph} (see Figure \ref{fig:three_reps}).


The label graph will be used to get an estimate of structural similarities. To generate the label graph, AMN substitutes each entity node's label with a generic entity token. This is intentional, as it reflects that entity labels have no inherent utility for producing matchings according to SMT. Then, each function and predicate node is assigned a randomly chosen generic label (from a fixed set of such labels) based off its arity and orderedness. Assignments are made consistently across the entire graph, e.g., \textit{every} instance of \texttt{MASS} in \textit{both} the base and target would be assigned the same generic replacement label. This substitution means the original label is not used in the matching process, which allows AMN to generalize to new symbols.

The label graph is not sufficient to produce representations that can be used for matching, as it represents a node by only label-based features which are shared amongst different nodes, 
an issue known as the \textit{type-token distinction} \cite{kahneman1992reviewing,wetzel2006types}. To contend with this, a signature graph is constructed that represents nodes in a way that respects object identity. To construct the signature graph, AMN replaces each distinct entity with a unique identifier (drawn from a fixed set of possible identifiers). It then assigns each function and predicate a new label based solely on its arity and orderedness, ignoring the original symbol. For instance, \texttt{ATTRACTS} and \texttt{REVOLVES-AROUND} would be assigned the same label as they are both ordered binary predicates.

As all input graphs will be DAGs, AMN uses two separate DAG LSTMs \cite{crouse2019improving} to embed the nodes of the label and signature graphs (equations detailed in Appendix 7.4
). Each node embedding is computed as a function of its complete set of dependencies in the original graph.
The set of label structure embeddings is written as $L_V = \{ l_v : v \in V \}$ and the set of signature embeddings is written as $S_V = \{ s_v : v \in V \}$. Before passing these embeddings to the next step, each element of $S_V$ is scaled to unit length, i.e. each $s_v$ becomes $s_v / \|s_v\|$, which gives our network an efficiently checkable criterion for whether or not two nodes are likely to be equal, i.e., when the dot product of two signature embeddings is 1.

\subsubsection{Correspondence Selector:} The graph embedding procedure yields two sets of node embeddings (label structure and signature embeddings) for the base and target. We utilize the set of embedding pairs for each node of $V_B$ and $V_T$, writing $l_v$ to denote the label structure embedding of node $v$ from $L_V$ and $s_v$ the signature embedding of node $v$ from $S_V$. We first define the set of unprocessed correspondences $\mathcal{C}^{(0)}$
\begin{alignat*}{2}
    &\hat{\mathcal{C}} &&= \{ \big< b, t \big> \in V_B \times V_T : \| l_b - l_t \| \leq \epsilon \} \\
&\mathcal{C}^{(0)} &&= \{ \big< \big[l_b; l_t; s_b; s_t\big], s_b, s_t \big> : \big< b, t \big> \in \hat{\mathcal{C}} \}
\end{alignat*}
where $[ \cdot ; \cdot ]$ denotes vector concatenation, $\epsilon$ is the tiered identicality threshold that governs how much the subgraphs rooted at two nodes may differ and still be considered for correspondence (in this work, we set $\epsilon = 1\mathrm{e}{-5}$).
The first element of each correspondence in $\mathcal{C}^{(0)}$, i.e., $h_c = \big[l_b; l_t; s_b; s_t\big]$, is then passed through an $N$-layered Transformer encoder (equations detailed in Appendix 7.4
) to produce a set of encoded correspondences $\mathcal{E} = \{ \big< h_c^{(N)}, s_b, s_t \big> \in \mathcal{C}^{(N)} \}$.

\begin{figure*}[t]
    \centering
    \includegraphics[width=\textwidth]{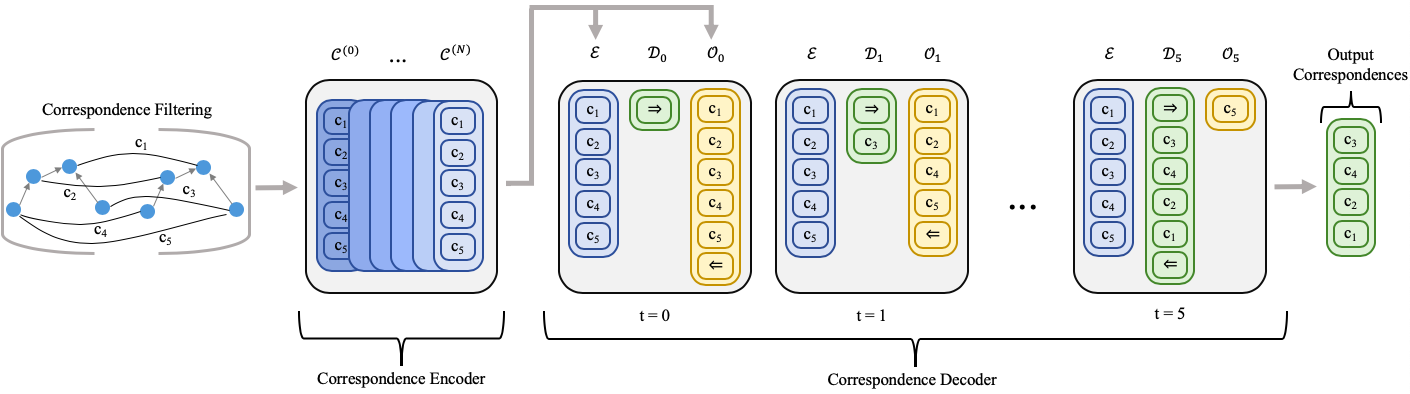}
    \caption{The correspondence selection process, where $\Rightarrow$ and $\Leftarrow$ are the start and stop tokens and $\mathcal{E}$, $\mathcal{D}_t$, and $\mathcal{O}_t$ are the sets of encoded, selected, and remaining correspondences}
    \label{fig:enc_dec}
\end{figure*}

The Transformer decoder selects a subset of correspondences that constitutes the best analogical match (see Figure~\ref{fig:enc_dec}). The attention-based transformations are only performed on the initial element of each tuple, i.e., $h_d$ in $\big<h_d, s_{b}, s_{t}\big>$. We let $\mathcal{D}_t$ be the processed set of all selected correspondences at timestep $t$ (after the $N$ attention layers) and $\mathcal{O}_t$ be the set of all remaining correspondences (with $\mathcal{D}_0 = \{ \texttt{START-TOK}\}$ and $\mathcal{O}_0 = \mathcal{E} \cup \{\texttt{END-TOK}\}$). The decoder generates compatibility scores $\alpha_{od}$ between each pair of elements, i.e., $\big<o, d\big> \in \mathcal{O}_t \times \mathcal{D}_t$. These are combined with the signature embedding similarities to produce a final compatibility $\pi_{od}$
\begin{equation*}
    \pi_{od} = \textrm{FFN}\big(\big[\tanh{(\alpha_{od})}; s_{b_o}^\top s_{b_d}; s_{t_o}^\top s_{t_d}\big]\big)
\end{equation*}
where FFN is a two layer feed-forward network with ELU activations \cite{clevert2015fast}.
Recall that the signature components, i.e. $s_b$ and $s_t$, were scaled to unit length. Thus, we would expect closeness in the original graph to be reflected by dot-product similarity and identicality to be indicated by a maximum value dot-product, i.e. $s_{b_o}^\top s_{b_d} = 1$ or $s_{t_o}^\top s_{t_d} = 1$. Once each pair has been scored, AMN selects an element of $\mathcal{O}_t$ to be added to $\mathcal{D}_{t + 1}$. For each $o \in \mathcal{O}_t$, we compute its value to be
\begin{equation*}
    v_{o} = \textrm{FFN}\big(\big[ \max_{d} \pi_{od}; \min_{d} \pi_{od}; \sum_{d}\dfrac{\pi_{od}}{|\mathcal{D}_t|}  \big]\big)
\end{equation*}
where FFN is a two layer feed-forward network with ELU activations. A softmax is applied to these scores and the highest valued element is added to $\mathcal{D}_{t + 1}$. The use of maximum, minimum, and average  is intended to let the network capture both individual and aggregate evidence. Individual evidence is given by a pairwise interaction between two correspondences (e.g., two correspondences that together violate the one-to-one constraint). Conversely, aggregate evidence is given by the interaction of a correspondence with everything selected thus far (e.g., a correspondence needed for several parallel connectivity constraints). When \texttt{END-TOK} is selected, the set of correspondences $M$ returned is the set of node pairs from $V_B$ and $V_T$ associated with elements in $\mathcal{D}$. 

\subsubsection{Candidate Inference Selector:} The output of the correspondence selector is a set of correspondences $M$. The candidate inferences associated with $M$ are drawn from the nodes of the base graph $V_B$ that were \emph{not} used in $M$. Let $V_{in}$ and $V_{out}$ be the subsets of $V_B$ that were / were not used in $M$, respectively. We first extract all signature embeddings for both sets, i.e., $\mathcal{S}_{in} = \{ s_b : b \in V_{in} \}$ and $\mathcal{S}_{out} = \{ s_b : b \in V_{out} \}$. In this module there are no Transformer components, with AMN operating directly on $\mathcal{S}_{in}$ and $\mathcal{S}_{out}$.

AMN will select elements from $\mathcal{S}_{out}$ to return. Like before, we let $\mathcal{D}_t$ be the set of all selected elements from $\mathcal{S}_{out}$ and $\mathcal{O}_t$ be the set of all remaining elements from $\mathcal{S}_{out}$ at timestep $t$. AMN computes compatibility scores between pairs of output options with candidate inference and previously selected nodes, i.e. $\alpha_{od}$ for each $\big<o, d\big> \in \mathcal{O}_t \times (\mathcal{D}_t \cup \mathcal{S}_{in})$. The compatibility scores are given by a simple single-headed attention computation (see Appendix 7.4
). Unlike the correspondence encoder-decoder, there are no other values to combine these scores with, so they are used directly to compute a value $v_o$ for each element of $\mathcal{O}_t$. AMN computes this value as
\begin{alignat*}{2}
&\alpha_{od}^\prime &&= \tanh{(\alpha_{od})} \\
&v_{o} &&= \textrm{FFN}\big(\big[ \max_{d} \alpha_{od}^\prime; \min_{d} \alpha_{od}^\prime; \sum_{d}\dfrac{\alpha_{od}^\prime}{|\mathcal{D}_t|}  \big]\big)
\end{alignat*}
A softmax is used and the highest valued element is added to $\mathcal{D}_{t + 1}$. Once the end token is selected, decoding stops and the set of nodes associated with elements in $\mathcal{D}$ is returned.

\subsubsection{Loss Function:} As both the correspondence and candidate inference components use a softmax, the loss function is categorical cross entropy. Teacher forcing is used to guide the decoder to select the correct choices during training. With $\mathcal{L}_{corr}$ the loss for correspondence selection and $\mathcal{L}_{ci}$ the loss for candidate inference selection, the final loss is given as $\mathcal{L} = \mathcal{L}_{corr} + \lambda \mathcal{L}_{ci}$ (with $\lambda$ a hyperparameter), which is minimized with Adam \cite{kingma2014adam}.

\subsection{Model Scoring}
\label{sec:tr_sc}

\subsubsection{Structural Match Scoring:} In order to avoid counting erroneous correspondence predictions towards the score of the output correspondences $M$, we first identify all correspondences that are either degenerate or violate the constraints of SMT. Degenerate correspondences are correspondences between constants that have no higher-order structural support in $M$ (i.e., if either has no parent that participates in a correspondence in $M$). To determine if a correspondence $\big<b, t\big>$ violates SMT, we check whether the subgraphs of the base and target rooted at $b$ and $t$ satisfy the one-to-one matching, parallel connectivity, and tiered identicality constraints (see Section \ref{sec:smt}). The check can be computed in time linear with the size of the corresponding subgraphs.
Let the valid subset of $M$ be $M_{val}$. A correspondence $m$ is considered a \emph{root correspondence} if there does not exist another correspondence $m^\prime$ such that $m^\prime \in M_{val}$ and a node in $m^\prime$ is an ancestor of a node in $m$. We define $M_{root} \subseteq M_{val}$ to be the set of all such root correspondences. For a correspondence $m = \big< b, t \big>$ in $M_{val}$, its score $s(m)$ is given as the size of the subgraph rooted at $b$ in the base. The structural match score for $M$ is then sum of scores for all correspondences in $M_{root}$, i.e., $s(M) = \sum_{m \in M_{root}} s(m)$. This repeatedly counts nodes that appear in the dependencies of multiple correspondences, which leads to higher scores for more interconnected matchings (in keeping with the systematicity preference of SMT).

\subsubsection{Structural Evaluation Maximization:} Dynamically assigning labels to each example allows AMN to handle never-before-seen symbols, but its inherent randomness can lead to significant variability in terms of outputs. AMN combats this by running each test problem $r$ times and returning the mapping $M = \argmax_{M_i} \sum_j J(M_i, M_j)$, where $J(M_i, M_j)$ is the Jaccard index (intersection over union) between the correspondence sets produced by the $i$-th and $j$-th runs. Intuitively, this is the run that shared the most correspondences with other runs and had the fewest unshared extra correspondences. 

\section{Experiments}

\subsection{Data Generation and Training}

AMN was trained on 100,000 synthetic analogy examples, with the hyperparameters used for AMN provided in Appendix 7.1 
(in the supplementary material). A single example consisted of base and target graphs, a set of correspondences, and a set of nodes from the base to be candidate inferences. Construction of synthetic examples begins with generating DAGs. Each DAG consists of a set of $k \in [2, 7]$ layers (with the particular $k$ for a graph chosen at random). Each node is assigned an arity $a$, with the maximum arity being $a = 3$. Nodes at layer $i$ can be connected to $a$ nodes from lower layers (i.e., layer $j$ with $j < i$) selected at random. Nodes with arity $a = 0$ are considered entities and nodes with non-zero arities (i.e., $a > 0$) are randomly assigned as predicates or functions and randomly designated as ordered or unordered.

To generate a training example, we first generate a set of random DAGs $C$, which will later become the correspondences. Next, we construct the base $B$ by generating graphs \emph{above} $C$. As each DAG is constructed in layers, this simply means that $C$ is considered the lowest layers of $B$. Likewise, for the target $T$ we build another set of graphs above $C$. The nodes of $C$ are thus shared with both $B$ and $T$. Each node of $C$ is duplicated, producing one node for $B$ and one node for $T$, and the resulting pair of nodes becomes a correspondence. Any element in $B$ that was an ancestor of a node from $C$ or a descendent of such an ancestor was considered a candidate inference. In Appendix 7.2 
we provide a  figure showing each  component of a training example. During training, each generated example was turned into a batch of 8 inputs by repeatedly running the encoding procedure (which dynamically assigns node labels) over the original base and target.

\begin{figure*}[t]
\renewcommand*{\arraystretch}{1.}
    \centering
    \includegraphics[width=0.75\textwidth]{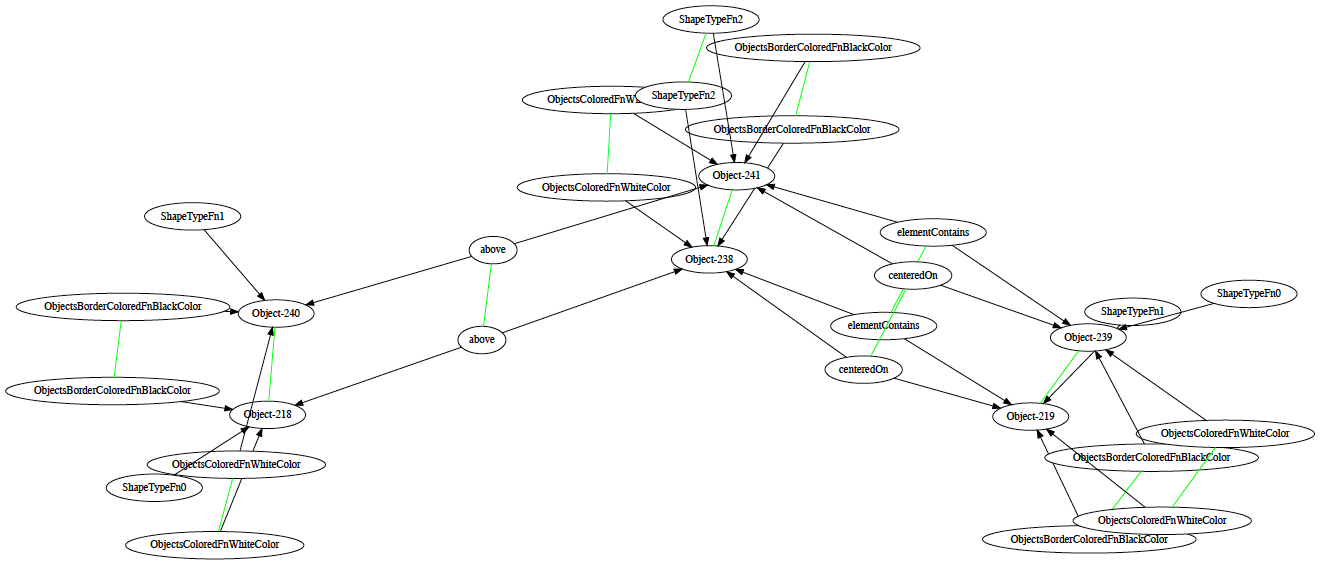}
    \caption{AMN output correspondences for an example from the Geometric Analogies domain}
    \label{fig:geo_amn_ex}
\end{figure*}

\subsection{Experimental Domains}

Though training was done with synthetic data, we evaluated the effectiveness of AMN on both synthetic data and data used in previous analogy experiments. The corpus of previous analogy examples was taken from the public release of SME\footnote{http://www.qrg.northwestern.edu/software/sme4/index.html}. Importantly, AMN was \emph{not} trained on the corpus of existing analogy examples (AMN never learned from a real-world analogy example). In fact, there was \emph{no} overlap between the symbols (i.e., entities, functions, and predicates) used in that corpus and the symbols used for the synthetic data. We briefly describe each of the domains AMN was evaluated on below (detailed descriptions can be found in \cite{forbus2017extending}).
\begin{enumerate}[leftmargin=*]
    \item \textit{Synthetic}: this domain consisted of 1000 examples generated with the same parameters as the training data (useful as a sanity check for AMN's performance).
    \item \textit{Visual Oddity}: this problem setting was initially proposed to explore cultural differences to geometric reasoning in \cite{dehaene2006core}. The work of \cite{lovett2011cultural} modeled the findings of the original experiment computationally with qualitative visual representations and analogy. We extracted 3405 analogical comparisons from the computational experiment.
    \item \textit{Moral Decision Making}: this domain was taken from \cite{dehghani2008moraldm}, which introduced a computational model of moral decision making that used SME to reason through moral dilemmas. From the works of \cite{dehghani2008moraldm,dehghani2008integrated}, we extracted 420 analogical comparisons.
    \item \textit{Geometric Analogies}: this domain is from one of the first computational analogy experiments \cite{evans1964program}. Each problem was an incomplete analogy of the form $A : B :: C : \ ?$, where each of $A$, $B$, and $C$ were manually encoded geometric figures and the goal was to select the  figure that best completed the analogy from an encoded set of possible answers. While in the original work all figures had to be manually encoded, in  \cite{lovett2009solving,lovett2012modeling} it was shown that the analogy problems could be solved with structure-mapping over automatic encodings (produced by the CogSketch system \cite{forbus2011cogsketch}). From that work we extracted 866 analogies.
\end{enumerate}

\subsection{Results and Discussion}


\begin{table*}[t]
\centering
\begin{subtable}{\textwidth}
\centering
\footnotesize
\setlength\tabcolsep{5pt}%
\begin{tabular}{ l | c || c | c c c | c c c }
\toprule 
 Domain & $r$ & Struct. Perf. & Larger & Equiv. & Err. Free & 1-to-1 Err. & PC Err. & Degen. Err. \\
\midrule
Synthetic & 1 & 0.713 & 0.000 & 0.313 & 0.346 & 0.007 & 0.102 & 0.020 \\
Synthetic & 16 & 0.952 & 0.001 & 0.683 & 0.695 & 0.005 & 0.020 & 0.011 \\\hline
Oddity & 1 & 0.774 & 0.061 & 0.404 & 0.484 & 0.153 & 0.225 & 0.000 \\
Oddity & 16 & 0.955 & 0.074 & 0.485 & 0.564 & 0.131 & 0.139 & 0.000 \\\hline
Moral DM & 1 & 0.610 & 0.014 & 0.021 & 0.093 & 0.002 & 0.170 & 0.030 \\
Moral DM & 16 & 0.958 & 0.081 & 0.164 & 0.329 & 0.000 & 0.041 & 0.016 \\\hline
Geometric & 1 & 0.871 & 0.064 & 0.533 & 0.649 & 0.039 & 0.116 & 0.000 \\
Geometric & 16 & 1.040 & 0.069 & 0.714 & 0.788 & 0.029 & 0.043 & 0.000 \\
\bottomrule
\end{tabular}
\caption{AMN correspondence prediction results for performance ratio (left), solution type rate (middle, $\uparrow$~better), and error rate (right, $\downarrow$~better)}
\label{res:exp_perf}
\end{subtable}
\vskip 0.1in
\begin{subtable}{\textwidth}
\centering
\footnotesize
\begin{tabular}{ l | c || c c c c c }
\toprule 
 Domain & $r$ & Avg. CI F1 & Avg. CI Prec. & Avg. CI Rec. & Avg. CI Acc. & Avg. CI Spec. \\
\midrule
Synthetic & 16 & 0.900 & 0.867 & 0.967 & 0.861 & 0.735 \\
Oddity & 16 & 0.992 & 0.995 & 0.994 & 0.991 & 0.911 \\
Moral DM & 16 & 0.899 & 0.834 & 0.985 & 0.832 & 0.439 \\
Geometric & 16 & 0.958 & 0.955 & 0.990 & 0.951 & 0.917 \\
\bottomrule
\end{tabular}
\caption{AMN candidate inference prediction results}
\label{res:exp_ci}
\end{subtable}
\caption{AMN experimental results}
\label{res:exp_all}
\end{table*}

Table \ref{res:exp_perf} shows the results for AMN across different values of $r$, where $r$ denotes the re-run hyperparameter detailed in Section~\ref{sec:tr_sc}. When evaluating on the synthetic data, the comparison set of correspondences was given by the data generator; whereas when evaluating on the three other analogy domains, the comparison set of correspondences was given by the output of SME. It is important to note that we are using SME as our stand-in for SMT (as it is the most widely accepted computational model of SMT). Thus, we do \emph{not} want significantly different results from SME in the correspondence selection experiments (e.g., substantially higher or lower structural evaluation scores). Matching SME's performance (i.e., not producing higher or lower values) gives evidence that we are modeling SMT.

In the Struct.~Perf. column, the numbers reflect the average across examples of the structural evaluation score of AMN divided by that of the comparison correspondence sets. For the other columns of Table \ref{res:exp_perf}, the numbers represent average fractions of examples or correspondences (e.g., 0.684 should be interpreted as 68.4\%). Candidate inference prediction performance was measured relative to the set of correspondences AMN generated, i.e., all candidate inferences were computed from the \emph{predicted} correspondences, and treated as the true positives. In many problems from the non-synthetic domains, every non-correspondence node was a candidate inference (which can lead to inflated precision and recall values). Thus, we also report the specificity (i.e., true negative rate) of AMN for \emph{only} problems with non-candidate inference nodes.

In addition to our main results, we also provide qualitative examples of AMN's outputs on real analogy problems and ablation studies for various aspects of AMN's design. Both the matching shown in Figure \ref{fig:geo_amn_ex} as well as the solar system analogy shown in Figure \ref{fig:rel_repr} were produced by AMN. Further examples of AMN's outputs can be found in Appendix 7.5. 
Ablation experiments regarding the impact of both the signature graph and unit normalization of signature embeddings (each detailed in Section \ref{sec:model_comps}) are given in Appendix 7.3.

\subsubsection{Analysis:} The left side of Table \ref{res:exp_perf} shows the average ratio of AMN's performance (labeled Struct.~Perf.), as measured by structural evaluation score, against the comparison method's performance (i.e., data generator correspondences or SME). As can be seen, AMN produced matches with structural evaluation scores at 95-104\% the level of SME on the non-synthetic domains, which indicates that it was finding similar structural matches. This is ideal as it shows that AMN matches SME's systematicity preference, and thus likely conforms fairly well to SMT in terms of systematicity.

The middle of Table \ref{res:exp_perf} gives us the best sense of how well AMN modeled SMT. We observe AMN's performance in terms of the proportion of \emph{larger}, \emph{equivalent}, and \emph{error-free} matches it produces (labeled Larger, Equiv., and Err.~Free, respectively). Error-free matches do not contain degenerate correspondences or SMT constraint violations, whereas equivalent and larger matches are both error-free and have the same / larger structural evaluation score as compared to gold set of correspondences. The Equiv. column provides the best indication that AMN could model SMT. It shows that $\gtrapprox 50\%$ of AMN's outputs were SMT-satisfying, error-free analogical matches with the \emph{exact same} structural score as SME (the lead computational model of SMT) in two of the non-synthetic analogy domains. 

The right side of Table \ref{res:exp_perf} shows the frequency of the different types of errors, including violations of the one-to-one / parallel connectivity constraints, and degenerate correspondences (labeled 1-to-1 Err., PC Err., and Degen. Err.). It shows that AMN had fairly low error rates across domains (except for Visual Oddity). Importantly, degenerate correspondences were very infrequent, which is significant because it verifies that AMN leveraged higher-order relational structure.


Table \ref{res:exp_ci} shows that AMN was fairly effective in predicting candidate inferences. The high accuracy (labeled Avg.~CI~Acc.) scores for both the Visual Oddity and Geometric Analogies domains indicate that AMN was able to capture the notion of structural support when determining candidate inferences. The non-zero specificity (labeled Avg.~CI~Spec.) results show that, while it more often classified nodes as candidate inferences, it was capable of distinguishing non-candidate inference nodes as well.


\section{Conclusions}

In this paper, we introduced the Analogical Matching Network, a neural approach that learned to produce analogies consistent with Structure-Mapping Theory. AMN was trained on completely synthetic data and was capable of performing well on a varied set of analogies drawn from previous work involving analogical reasoning. AMN demonstrated renaming invariance, structural sensitivity, and the ability to find solutions in a combinatorial search space, all of which are key properties of symbolic reasoners and are known to be important to human reasoning.


\bibliographystyle{aaai}
\bibliography{references}

\section{Appendix}

\subsection{Model Details}
\label{sec:hyperparams}
In the DAG LSTM, the node embeddings were 32-dimensional vectors and the edge embeddings were 16-dimensional vectors. For all Transformer components, our model used multi-headed attention with 2 attention layers each having 4 heads. In each multi-headed attention layer, the query and key vectors were projected to 128-dimensional vectors. The feed forward networks used in the Transformer components had one hidden layer with a dimensionality twice that of the input vector size. The feed forward networks used to compute the values in the correspondence selector used two 64-dimensional hidden layers. The $\lambda$ parameter applied to the candidate inference loss $\mathcal{L}_{ci}$ was set to $\lambda = 0.1$ in our experiments. The models were constructed with the Pytorch \cite{paszke2019pytorch} library.

\subsection{Training Data Generation}
\label{sec:tr_data_gen}

\begin{figure*}[t]
    \centering
    \includegraphics[width=\textwidth]{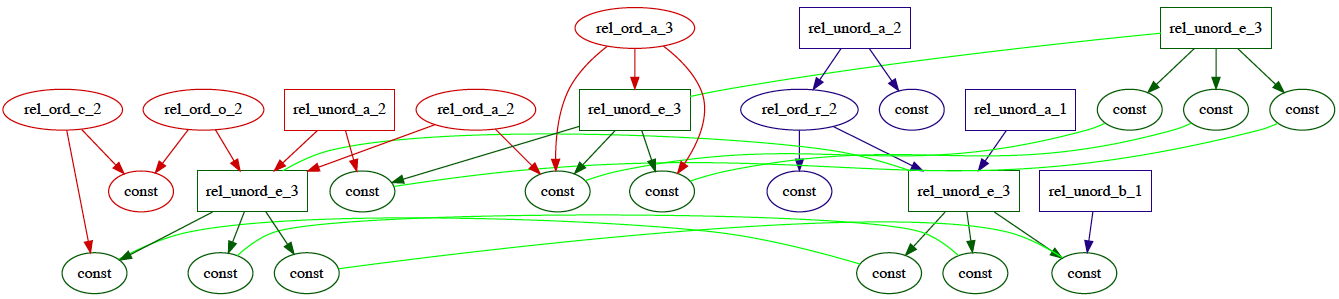}
    \caption{Synthetic example with a base (red), target (blue), and shared subgraphs (green)}
    \label{fig:rand_ex}
\end{figure*}

In Figure \ref{fig:rand_ex}, the dark green nodes indicate the initial random graphs $C$ \emph{after} being copied into the base and target. The red and blue nodes show the graphs built around $B$ and $T$. The light green edges indicate the gold set of correspondences generated from $C$.
On average, each example consisted of 26.9 expressions and 14.3 entities in the base (41.2 distinct items in total), 27.0 expressions and 14.3 entities in the target (41.3 distinct items in total), and 26.8 correspondences. 

\subsection{Additional Experiments}
\label{sec:ablations}
\subsubsection{Unit Normalization for Signature Embeddings:}

In Section \ref{sec:model_comps}, we described how signature embeddings were scaled to unit length to provide a simple criterion for whether two nodes were likely the same node (i.e., they have a dot product of 1). Intuitively, this feature would be most important for allowing AMN to follow SMT's one-to-one constraint, as it gives AMN the ability to determine which nodes have already been selected for correspondence. To measure the importance of this feature, we performed a simple experiment where we did \emph{not} scale the signature embeddings to unit length (keeping all other components of AMN the same). We retrained AMN following the same training methodology as before, and tested AMN on the synthetic domain.

Interestingly, we found that performance in all categories (not just conformance to SMT's one-to-one constraint) became significantly worse. The structural performance of AMN dropped from 0.948 to 0.750, indicating that systematicity was impacted. The fraction of problems that were equivalent to the gold standard correspondence set (i.e., no SMT errors and the same structural evaluation score as the gold standard) dropped from 0.671 to 0.278. In terms of errors, the percent of correspondences that violated one-to-one increased from 0.6\% to 1.6\% and those violating parallel connectivity increased from 2.1\% to 12.0\%. Degenerate errors remained about the same, increasing from 0.9\% to 1.2\%, likely reflecting that the dot product of two signature embeddings still incorporates their shared descendants.

\subsubsection{Value of the Signature Graph:}

Given that label graph captures almost all of the graph structure, it is natural to question whether the signature graph is necessary for producing SMT-conforming matchings. To determine the value of the signature graph, we performed an experiment where we completely excised the signature embeddings from AMN, leaving only the label graph for correspondence and candidate inference selection. We retrained this ablated version of AMN with the standard training methodology and tested it on the synthetic set of analogy problems.

Without the signature graph, AMN's performance plummeted in all categories. The one-to-one error rate increased from 0.6\% to 92.4\% and the parallel connectivity error rate increased from 1.2\% to 99.4\%. Consequently, the number of error free matches dropped to 0. This matches our intuitions, that without a distinction between the labels of objects and the objects themselves, AMN is incapable of modeling SMT.

\subsection{Background}
\subsubsection{DAG LSTMs:}
\label{sec:dag_lstm}

DAG LSTMs extend Tree LSTMs \cite{tai2015improved} to DAG-structured data. As with Tree LSTMs, DAG LSTMs compute each node embedding as the aggregated information of all their immediate predecessors (the equations for the DAG LSTM are identical to those of the Tree LSTM). The difference between the two is that DAG LSTMs stage the computation of a node's embedding based on the order given by a topological sort of the input graph. Batching of computations is done by grouping together updates of independent nodes (where two nodes are independent if they are neither ancestors nor predecessors of one another). As in \cite{crouse2019improving}, for a node, $v$, its initial node embedding, $s_v$, is assigned based on its label and arity. The DAG LSTM then computes the final embedding $h_v$ to be
\begin{alignat*}{2}
&i_v &&= \sigma \big( W_i s_v + \sum_{w \in \mathcal{P}(v)} U_i^{(e_{vw})} h_w + b_i \big) \\
&o_v &&= \sigma \big( W_o s_v + \sum_{w \in \mathcal{P}(v)} U_o^{(e_{vw})} h_w + b_o \big) \\
&\hat{c}_v &&= \tanh{ \big( W_c s_v + \sum_{w \in \mathcal{P}(v)} U_c^{(e_{vw})} h_w + b_c \big) } \\
&f_{vw} &&= \sigma \big( W_f s_v + U_f^{(e_{vw})} h_w + b_f \big) \\
&c_{v} &&= i_v \odot \hat{c}_v + \sum_{w \in \mathcal{P}(v)} f_{vw} \odot c_{w} \\
&h_{v} &&= o_v \odot \tanh{\big(c_v\big)}
\end{alignat*}
where $\odot$ is element-wise multiplication, $\sigma$ is the sigmoid function, $\mathcal{P}$ is the predecessor function that returns the arguments for a node, $U_i^{(e_{vw})}$, $U_o^{(e_{vw})}$, $U_c^{(e_{vw})}$, and $U_f^{(e_{vw})}$ are learned matrices per edge type. $i$ and $o$ represent input and output gates, $c$ and $\hat{c}$ are memory cells, and $f$ is a forget gate.

\subsubsection{Multi-Headed Attention:}
\label{sec:mha}

The multi-headed attention (MHA) mechanism of \cite{vaswani2017attention} is used in our work to compare correspondences against one another. In this work, MHA is given two inputs, a query vector $q$ and a list of key vectors to compare the query vector against $\big< k_1, \ldots, k_n\big>$. In $N$-headed attention, $N$ separate attention transformations are computed. For transformation $i$ we have
\begin{gather*}
    \hat{q}_{i} = W^{(q)}_{i} q, k_{ij} = W^{(k)}_{i} k_{j}, v_{ij} = W^{(v)}_{i} k_j \\
    w_{ij} = \dfrac{\hat{q}_{i}^\top k_{ij}}{\sqrt{b_{\hat{q}}}} \\
    \alpha_{ij} = \dfrac{\exp{(w_{ij})}}{\sum_{j^\prime}\exp(w_{ij^\prime})} \\
    q_{i} = \sum_j \alpha_{ij} v_{ij}
\end{gather*}
where each of $W^{(q)}_{i}$, $W^{(k)}_{i}$, and $W^{(v)}_{i}$ are learned matrices and $b_{\hat{q}}$ is the dimensionality of $\hat{q}_i$. The final output vector $q^\prime$ for input $q$ is then given as a combination of its $N$ transformations
\begin{gather*}
    q^\prime = \sum_{i = 1}^{N} W^{(o)}_i q_i
\end{gather*}
where each $W^{(o)}_i$ is a distinct learned matrix for each $i$. In implementation, the comparisons of query and key vectors are batched together and performed as efficient matrix multiplications.

\subsubsection{Transformer Encoder-Decoder:}
\label{sec:enc_dec}
The Transformer-based encoder-decoder is given two inputs, a comparison set $\mathcal{C}$ and an output set $\mathcal{O}$. At a high level, $\mathcal{C}$ will be encoded into a new set $\mathcal{E}$, which will inform a selection process that picks elements of $\mathcal{O}$ to return. In the context of pointer networks, the set $\mathcal{O}$ begins as the encoded input set, i.e., $\mathcal{O} = \mathcal{E}$.

\paragraph{Encoder:} First, the elements of $\mathcal{C}$, i.e. $h_c \in \mathcal{C}$, are passed through $N$ layers of an attention-based transformation. For element $h_c$ in the $i$-th layer (i.e., $h_c^{(i-1)}$) this is performed as follows
\begin{alignat*}{2}
&\hat{h}_c &&= \textrm{LN}\big(h_c^{(i - 1)} + \textrm{MHA}^{(i)}_{\mathcal{C}}\big(h_c^{(i - 1)}, \big<h_1^{(i - 1)}, \ldots, h_j^{(i - 1)}\big>\big)\big) \\
&h_c^{(i)} &&= \textrm{LN}\big(\hat{h}_c + \textrm{FFN}^{(i)}\big(\hat{h}_c\big)\big)
\end{alignat*}
where LN denotes the use of layer normalization \cite{ba2016layer}, $\textrm{MHA}^{(i)}_{\mathcal{C}}$ (Appendix \ref{sec:mha}) denotes the use of self multi-headed attention for layer $i$ (i.e., attention between $h^{(i)}_c$ and the other elements of $\mathcal{C}^{(i - 1)}$), and $\textrm{FFN}^{(i)}$ is a two-layer feed-forward neural network with ELU \cite{clevert2015fast} activations. After $N$ layers of processing, the set of encoded inputs $\mathcal{E}$ is given by $\mathcal{E} = \mathcal{C}^{(N)}$

\paragraph{Decoder:} With encoded comparison elements $\mathcal{E}$ and a set of potential outputs $\mathcal{O}$, the objective of the decoder is to use $\mathcal{E}$ to inform the selection of some subset of output options $\mathcal{D} \subseteq \mathcal{O}$ to return. Decoding happens sequentially; at each timestep $t \in \{1, \ldots, n\}$ the decoder selects an element from $\mathcal{O} \cup \{ \texttt{END-TOK} \}$ (where \texttt{END-TOK} is a learned triple) to add to $\mathcal{D}$. If \texttt{END-TOK} is chosen, the decoding procedure stops and $\mathcal{D}$ is returned. 

Let $\mathcal{D}_t$ be the set of elements that have been selected by timestep $t$ and $\mathcal{O}_t$ be the remaining unselected elements at timetstep $t$. First, $\mathcal{D}_t$ is processed with an $N$-layered attention-based transformation. For an element $h_d^{(i - 1)}$ this is given by
\begin{alignat*}{2}
&\acute{h}_d &&= \textrm{LN}\big(h_d^{(i - 1)} + \textrm{MHA}^{(i)}_{\mathcal{D}}\big(h_d^{(i - 1)}, \big<h_1^{(i - 1)}, \ldots, h_j^{(i - 1)}\big>\big)\big) \\
&\hat{h}_d &&= \textrm{LN}\big(\acute{h}_d + \textrm{MHA}^{(i)}_{\mathcal{E}}\big(\acute{h}_d, \big<h_1^{(i - 1)}, \ldots, h_l^{(i - 1)}\big>\big)\big) \\
&h_d^{(i)} &&= \textrm{LN}\big(\hat{h}_d + \textrm{FFN}^{(i)}\big(\hat{h}_d\big)\big)
\end{alignat*}
where $\textrm{MHA}^{(i)}_{\mathcal{D}}$ denotes the use of self multi-headed attention, $\textrm{MHA}^{(i)}_{\mathcal{E}}$ denotes the use of multi-headed attention against elements of $\mathcal{E}$, and $\textrm{FFN}^{(i)}$ is a two-layer feed-forward neural network with ELU activations. We will consider the already selected outputs to be the transformed selected outputs, i.e., $\mathcal{D}_t = \mathcal{D}_t^{(N)}$. For a pair, $\big<h_o, h_d\big> \in \mathcal{O}_t \times \mathcal{D}_t$, we compute their compatibility as $\alpha_{od}$
\begin{gather*}
    q_{od} = W_q h_d^{(n)}, k_{od} = W_k h_o \\
    \alpha_{do} = \dfrac{q_{od}^\top k_{od}}{\sqrt{b_o}}
\end{gather*}
where $W_q$ and $W_k$ are learned matrices, $b_o$ is the dimensionality of $h_o$, and FFN is a two layer feed-forward network with ELU activations. This defines a matrix $H \in \mathbb{R}^{|\mathcal{O}_t| \times |\mathcal{D}_t|}$ of compatibility scores. One can then apply some operation (e.g., max pooling) to produce a vector of values $v_t \in \mathbb{R}^{|\mathcal{O}_t|}$ which can be fed into a softmax to produce a distribution over options from $\mathcal{O}_t$. The highest probability element $\delta^*$ from the distribution is then added to the set of selected outputs, i.e., $\mathcal{D} = \mathcal{D}_t \cup \{ \delta^* \}$.

\subsection{AMN Example Outputs}
\label{ex_amn_outputs}

For the outputs from the non-synthetic domains (all but the first figure), only small subgraphs of the original graphs are shown (the original graphs were too large to be displayed)

\begin{figure*}[t]
    \centering
    \includegraphics[width=\textwidth]{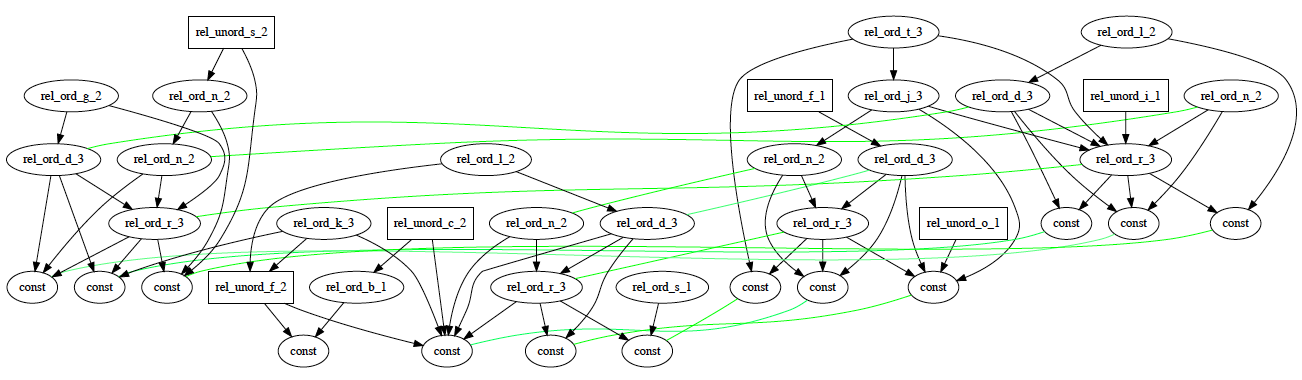}
    \caption{AMN output for an example from the Synthetic domain}
    \label{fig:synth_amn_ex}
\end{figure*}

\begin{figure*}[t]
    \centering
    \includegraphics[width=\textwidth]{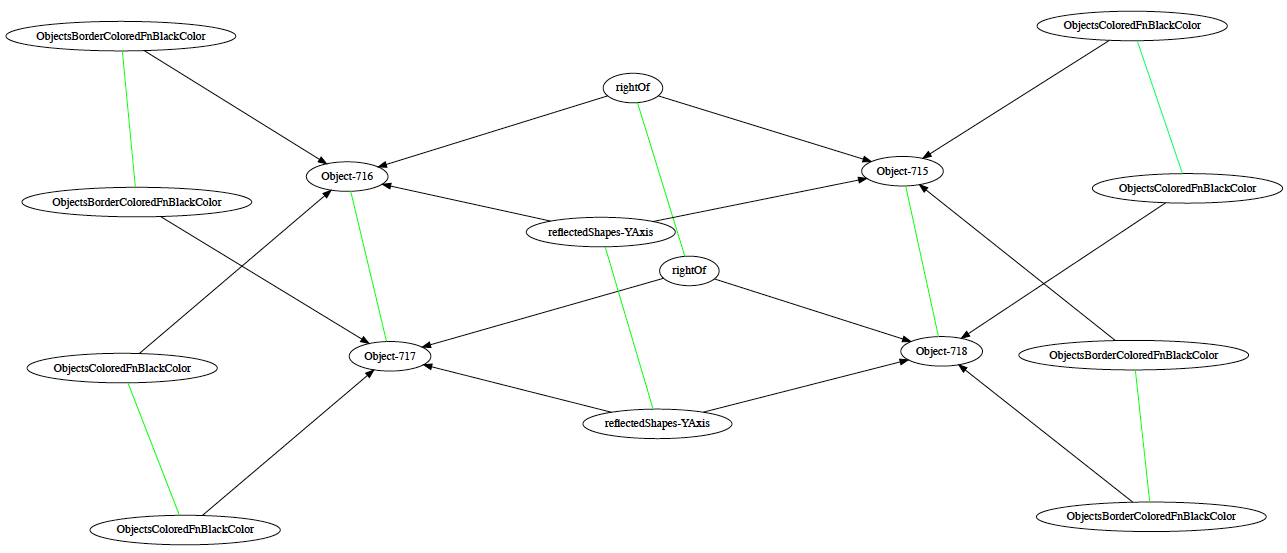}
    \caption{AMN output for an example from the Visual Oddity domain}
    \label{fig:odd_amn_ex}
\end{figure*}

\begin{figure*}[t]
    \centering
    \includegraphics[width=\textwidth]{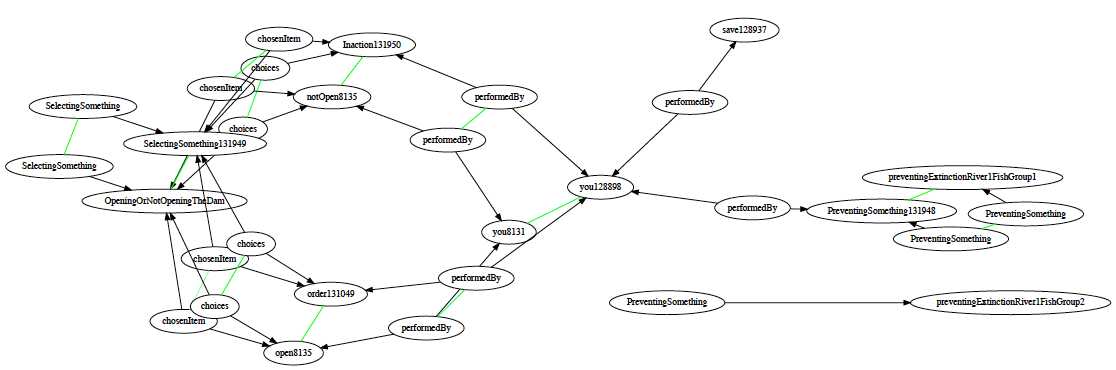}
    \caption{AMN output for an example from the Moral Decision Making domain}
    \label{fig:mdm_amn_ex}
\end{figure*}

\end{document}